\documentclass{article}

\usepackage{arxiv}

\usepackage[utf8]{inputenc} 
\usepackage[T1]{fontenc}    

\usepackage{hyperref}
\hypersetup{hidelinks}

\usepackage{url}            
\usepackage{booktabs}       
\usepackage{amsfonts}       
\usepackage{nicefrac}       
\usepackage{microtype}      
\usepackage{lipsum}
\usepackage{graphicx}
\usepackage{multirow}
\usepackage[numbers,square]{natbib}
\usepackage{doi}
\usepackage{amsmath}
\usepackage[bottom]{footmisc}
\usepackage{footnote}
\usepackage{natbib}
\usepackage{lipsum}
\usepackage{algorithm}
\usepackage{algpseudocode}
\usepackage[inline,shortlabels]{enumitem}
\usepackage{subcaption}
\usepackage{amssymb}
\usepackage{textcomp}
\usepackage{xcolor}
\usepackage{makecell}
\usepackage{float}
\usepackage{arydshln}

\title{CSG: A Context-Semantic Guided Diffusion Approach in De Novo Musculoskeletal Ultrasound Image Generation}

\date{}


\author{  \href{https://orcid.org/0009-0009-5635-6763}{\includegraphics[scale=0.06]{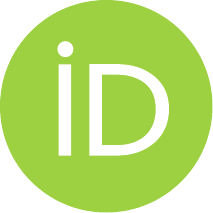}\hspace{1mm}Elay Dahan$^{1, \dag}$} \\
	AI/ML Research, GE Healthcare\\
    \And
\href{https://orcid.org/0000-0003-3741-9166}{\includegraphics[scale=0.06]{orcid.pdf}\hspace{1mm}Hedda Cohen Indelman$^{1, \dag}$} \\
    AI/ML Research, GE Healthcare\\
		\And
  \href{https://orcid.org/0009-0009-0370-5448}{\includegraphics[scale=0.06]{orcid.pdf}\hspace{1mm}Angeles M. Perez-Agosto$^{2}$}\\
    Clinical Insights \& Development, GE Healthcare\\
    		\And
  \href{https://orcid.org/0009-0000-5542-0035}{\includegraphics[scale=0.06]{orcid.pdf}\hspace{1mm}Carmit Shiran$^{3}$}\\
    Clinical Insights \& Development, GE Healthcare\\
	    \And
     \href{https://orcid.org/0009-0000-5542-0035}{\includegraphics[scale=0.06]{orcid.pdf}\hspace{1mm}Gopal Avinash$^{4}$} \\
	AI/ML Research, GE Healthcare\\
        \And
     \href{https://orcid.org/0009-0009-0263-6102}{\includegraphics[scale=0.06]{orcid.pdf}\hspace{1mm}Doron Shaked$^{1}$} \\
	AI/ML Research, GE Healthcare\\
        \And
     \href{https://orcid.org/0000-0002-0939-3379}{\includegraphics[scale=0.06]{orcid.pdf}\hspace{1mm}Nati Daniel$^{1,\dag,}$\thanks{Corresponding author, e-mail: \href{mailto:nati.daniel@gehealthcare.com}{nati.daniel@gehealthcare.com}. $\dag$These authors have contributed equally to this work. $^{1}$Dept. of AI/ML Research, Science \& Technology Organization, GE Healthcare, Haifa, Israel. $^{2}$Dept. of Clinical Applications, Point of Care Ultrasound \& Handheld, Texas, USA. $^{3}$Dept. of Clinical Applications, Point of Care Ultrasound \& Handheld, Wisconsin, USA. $^{3}$Dept. of AI/ML Research, Science \& Technology Organization, GEHC, San Ramon, USA.
     }} \\
	AI/ML Research, GE Healthcare
}

\renewcommand{\headeright}{Dahan E et al.}
\renewcommand{\shorttitle}{CSG: A Context-Semantic Guided Diffusion Approach}

\begin{document}

\maketitle

\begin{abstract}
The use of synthetic images in medical imaging Artificial Intelligence (AI) solutions has been shown to be beneficial in addressing the limited availability of diverse, unbiased, and representative data. Despite the extensive use of synthetic image generation methods, controlling the semantics variability and context details remains challenging, limiting their effectiveness in producing diverse and representative medical image datasets. In this work, we introduce a scalable semantic and context-conditioned generative model, coined CSG (Context-Semantic Guidance). This dual conditioning approach allows for comprehensive control over both structure and appearance, advancing the synthesis of realistic and diverse ultrasound images. We demonstrate the ability of CSG to generate findings (pathological anomalies) in musculoskeletal (MSK) ultrasound images. Moreover, we test the quality of the synthetic images using a three-fold validation protocol. The results show that the synthetic images generated by CSG improve the performance of semantic segmentation models, exhibit enhanced similarity to real images compared to the baseline methods, and are undistinguishable from real images according to a Turing test. Furthermore, we demonstrate an extension of the CSG that allows enhancing the variability space of images by synthetically generating augmentations of anatomical geometries and textures.
\end{abstract}

\keywords{Deep Learning, Stable Diffusion, Image Generation, Medical Imaging, Large Language Model.}
\section{INTRODUCTION} \label{intro}
\begin{figure}[ht]
\centering
  \includegraphics[width=0.8\linewidth]{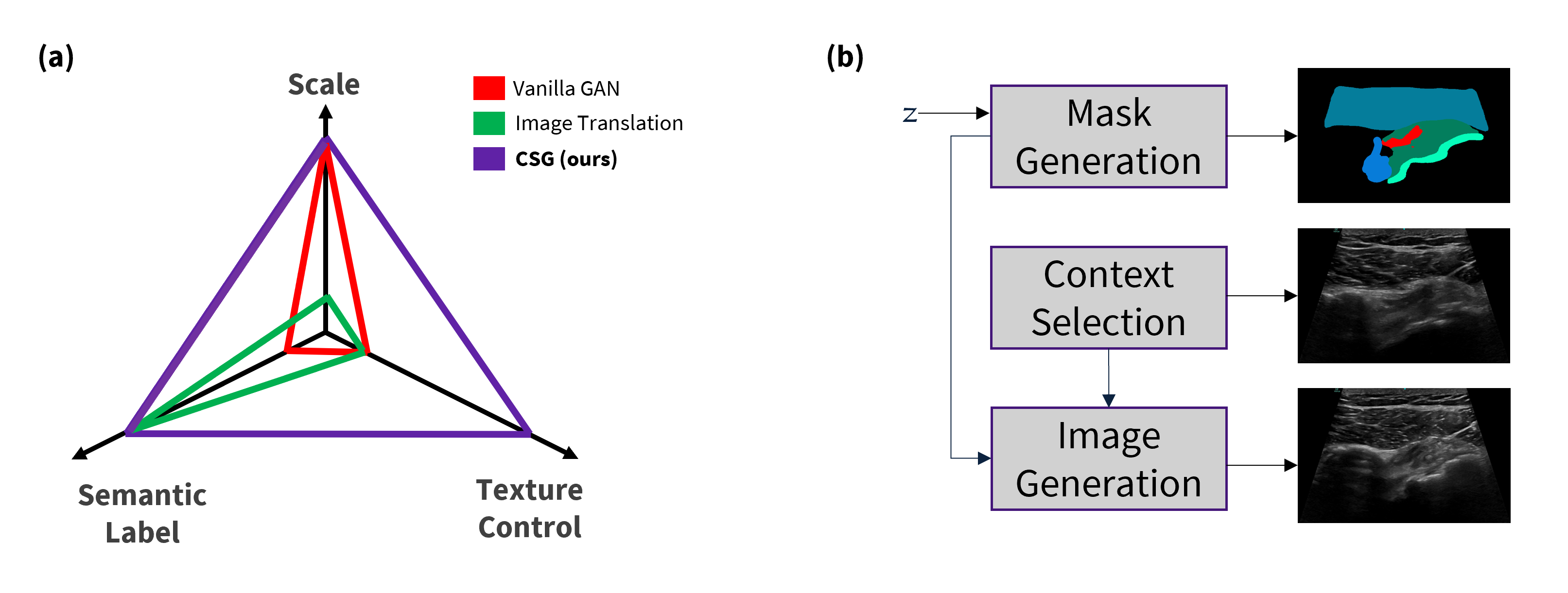}
    \caption{(a) Depiction of the differences between Image Translation GANs and Vanilla GANs, where the former creates synthetic images based on semantic labels. However, scalability is constrained when semantic labels are limited. In contrast, Vanilla GANs lack semantic information but offer the capability to generate an unlimited array of synthetic images. Addressing these challenges, our solution (CSG) is proposed. (b) CSG employs a triple-phase generative system. The initial step involves generating semantic masks of MSK labels using a fine-tuned StyleGAN architecture from a noise prior $z$. Subsequently, contextually similar images are selected using a neural algorithm of artistic style. Finally, the generated masks and contexts undergo processing through a paired image translation diffusion model to yield the synthetic ultrasound image. This approach harmonizes the advantages of semantic and context guidance for an unlimited and unbiased image generation.}
  \label{f:Fig0}
\end{figure}
Ultrasound is a widely used medical imaging modality for visualizing a diverse range of organs and tissues. It is frequently preferred due to its non-radiative, non-invasive characteristics, ease of use, rapid imaging capabilities, and relatively low cost. 
As the adoption of digital medical images became widespread, the use of AI methods increased in the field of ultrasound imaging \cite{tang2024development,katakis2023muscle,marzola2021deep,indelman2024semanticsegmentationrefinerultrasound}, and biomedical applications in general \cite{yasaka2018deep,chartrand2017deep,suzuki2017overview, czyzewski2021machine,daniel2022deep,larey2022fron}. However, despite the significant potential for AI in medical imaging, its demand for a large amount of data is an inherent limitation \cite{serag2019translational_46, tizhoosh2018artificial_47}. 
Moreover, in medical imaging, the identification of findings exhibiting potentially pathological anomalies presents a significant challenge, as these areas often lack clear boundaries, consistent geometry, or typical positioning. Furthermore, findings are typically rare, which exacerbates the difficulty of training diagnostic support models in scenarios with limited data.

Thus, a new perspective has emerged recently that underscores the equal significance of data in the AI modeling process. This emphasis is particularly crucial in instances near the decision boundaries, rare examples, and biased data. 
Addressing the shortage of diverse and comprehensive datasets, the integration of synthetic images emerged as a promising solution and has received growing interest in recent years, in the field of medical image in particular \cite{liang2022sketch,cao2023multi, che2021realistic, jiao2020self, dalmaz2022resvit}. 

Several prominent synthetic image generation methods have gained considerable traction. The widely recognized Generative Adversarial Networks (GANs) \cite{goodfellow2020generative} method allows generating samples that closely emulate an underlying data distribution based on adversarial training. GANs have proven to be instrumental in various image generation tasks \cite{radford2015unsupervised, guibas2017synthetic}, and may be conditioned to generate images from labels or semantic maps \cite{mirza2014conditional}.
Denoising Diffusion Probabilistic Models (DDPMs) \cite{sohl2015deep} introduce a distinctive method by utilizing a step-by-step denoising process. Recently, \cite{rombach2022high} suggested a new type of DDPMs, coined Latent Diffusion Models (LDM's), that perform the forward and reverse process in the latent space of a Variational Auto Encoder (VAE) \cite{kingma2022autoencodingvariationalbayes}, instead of in the image domain. LDM have shown to provide similar results to DDPM's with significantly lower computational requirements.
Image Translation Models (ITMs) \cite{cyclegan_paper, wang2018high, rombach2022high, daniel2023between, armanious2020medgan} transform images from one domain to another, learning mappings that preserve key content based on an input condition. These models excel at tasks of converting sketches to realistic images, changing seasons in photos, or enhancing resolution. In the context of synthetic medical image generation, translation models from the segmentation mask domain to the image domain are especially of interest. These models allow for controllable data generation and thus are of great value for synthetic medical images.

Despite the extensive use of synthetic image generation methods, a notable limitation remains in the generation of images with spatial variability \cite{yi2019generative}. As such, though GANs and DDPMs can be scaled through their sampling processes, their ability to control generative semantics is restricted. Our research addresses this by exploring semantic mask guidance to address the challenges in ultrasound image synthesis. However, a second issue emerges: while semantic masks offer control of spatial features, they fail to offer control of textual details, limiting their effectiveness in producing diverse and representative datasets.

To address these gaps, our study introduces an innovative dual conditioning of a state-of-the-art conditional latent diffusion model \cite{rombach2022high}. Our solution, illustrated in Fig. \ref{f:Fig0}, integrates semantic guidance to control the anatomical structure, ensuring accurate geometry in the generated images. Moreover, a contextual conditioning guide rich texture details in the image generation output. This dual conditioning approach allows for comprehensive control over both structure and appearance, advancing the synthesis of realistic and diverse ultrasound images. 

The paper is organized as follows: 
Our method is presented in Section \ref{csg_method}, and \ref{subsec:scalingCSG} and the experimental setup is presented in Section \ref{subsec:experimental}. 
Section \ref{results} presents the results and ablation studies on a discontinuity in tendon fiber (DITF) pathology finding task in a musculoskeletal ultrasound (MSK) dataset, and the conclusions are presented in Section \ref{conclusion}.

\section{MATERIALS AND METHODS} \label{methods_and_materials}
\subsection{The CSG method}\label{csg_method}
In this subsection, we present the building blocks of our method: a context selection algorithm that pairs each semantic mask with textual features to construct clinically valid guidance, and a dual-conditioning image generation model that accepts both semantic and texture guidance. 

Finally, extensions are introduced to allow scaling the image generation process both by generating semantic masks, and enhancing the image variability space by allowing editing of the geometry and texture guidance. These extensions may be applied separately or jointly.

\begin{figure}[t]
\centering
\vspace{9pt}
\includegraphics[width=0.5\linewidth]{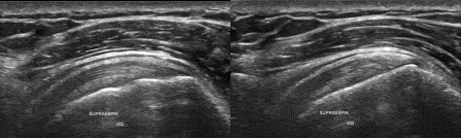}

\vspace{1pt}
\includegraphics[width=0.5\linewidth]{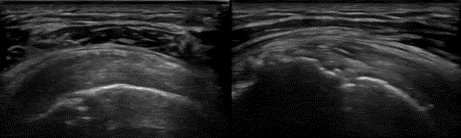}
        \caption{Examples of a query and contextual similar image pairs. On the left: a query image. On the right: The most visually similar image in the dataset based on texture style selected by our context selection method.}
        \label{f:f2}
\end{figure}

\subsubsection{Context selection}\label{CSG_context}
A context conditioning is introduced to control the texture properties of the output images. The context condition is an image with visual properties similar to those of the target output image. Following \cite{gatys2015neural}, we use the encoding of certain layers of a pre-trained CNN to extract textual features from images. Thus, for each image in our dataset, we calculate the textual features as described in \cite{gatys2015neural}. Then, for each image (denoted as prompt image) in the dataset ,we find the most similar non-equal feature vector by computing the Mean Squared Error (MSE) between these two features.
By that, we were able to find a visually similar image for any given image in the dataset.
We then constructed a dataset of two input images of semantic mask and context image paired with the ground-truth ultrasound image.

The results in Fig.~\ref{f:f2} show a query image (left) and the most visually similar image (right) in the dataset based on texture style. These results demonstrate that the images have visually a similar context. Indeed, the selected images are often from the same ultrasound clip (e.g. top row).

\subsubsection{Image generation}\label{CSG_image}
Our image-based generation is based on the conditional latent diffusion model \cite{rombach2022high}, mainly used for text-to-image translation based on the stable diffusion model. Our method aims to generate an image conditioned on a semantic mask and context. This involves adjusting the input of the denoising U-Net \cite{unet_2015} to accommodate two images. Furthermore, we adapt the double conditioning classifier-free guidance \cite{brooks2022instructpix2pix}, such that the modified sampling score estimation is defined in Eq. (\ref{eq:of}).
\begin{align}
\widetilde{e_{\theta}}(z_{t},c_{S},c_{C}) &= \phi + s_{S} \cdot (e_{\theta}(z_{t},c_{S}, \emptyset)- \phi) \label{eq:of} \\
    &+ s_{C} \cdot (e_{\theta}(z_{t},c_{S},c_{C})-e_{\theta}(z_{t},c_{S}, \emptyset)) \nonumber,
\end{align}
where $\phi$ denotes $e_{\theta}(z_{t}, \emptyset, \emptyset)$, $C_{C}, C_{S}$ indicates the context and semantic guidance image latent vector respectively. $s_{S}, s_{C}$ denote the guidance scale of the semantic image and context image, respectively, and are set to
$s_{S} = 1.5$ and $s_{C} = 2.5$ in all experiments unless otherwise stated. We follow previous work \cite{brooks2022instructpix2pix}, and randomly set $C_{C} = \emptyset$ or $C_{S} =\emptyset$ for $5\%$ of examples, and both $C_{C} = \emptyset$ and $C_{S} = \text{\o}$ for $5\%$ of examples.
Thus, with semantic guidance, control over the anatomical structure (geometry) of the output image is achieved. Further, to enhance the diversity of generated samples, we incorporated context (texture) guidance. 

\begin{figure}[t]
\centering
  \includegraphics[width=0.8\linewidth]{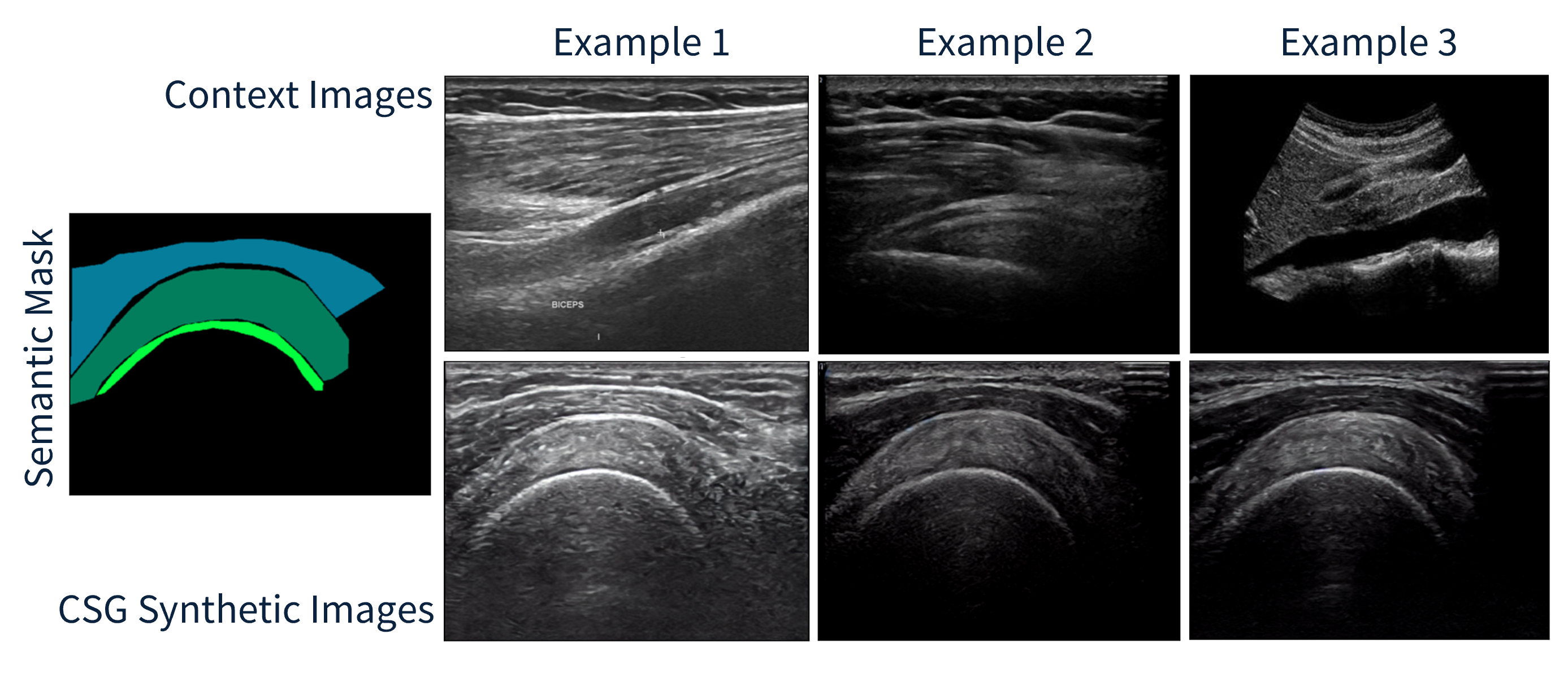}
    \caption{Example synthetic ultrasound images generated by our CSG method. A semantic mask (on the left) and context image (top row) guide the synthetic image generation of our method (bottom row). Label mapping: Blue steel - muscle, Green – tendon, Light green- bone, Black - background.}
  \label{f:f1}
\end{figure}

During inference, our model maps a pair of a semantic map and texture to an ultrasound image that aligns with both the geometry of the semantic map and the texture of the context image. Examples in Fig.~\ref{f:f1} show that for a geometry semantic mask, and different context images (styles) as guidance, our method's generated ultrasound images are noticeably highly correlated with the conditioned context images, even when the condition images are not from the dataset (e.g. the abdomen context image in the third column). Additionally, the outputs spatially match the input mask's geometry.

\subsection{Scaling the CSG method}\label{subsec:scalingCSG}
Training the CSG model requires pairs of images and corresponding semantic masks. However, due to the costly data acquisition and labeling task, the often limited training set impedes the performance and generability of such models. 
An enhanced CSG method addresses the challenges of scalability and variability in MSK diagnosis by integrating an unbiased and unlimited semantic mask generative model, as described in \cite{larey2023depas, daniel2024priorpath}.

\subsubsection{Mask generation}\label{CSG_mask}
Our end-to-end solution incorporates a mask generator network. Using StyleGAN \cite{karras2020training}, a generative model pretrained on the BRECAHAD dataset \cite{aksac2019brecahad}, is fine-tuned on MSK mask images. Given our primary objective of pathology finding, we implemented a filtering mechanism to exclude generated masks that lack significant pathologies.

\subsubsection{Extending the variability space}
We present an extension of the CSG that allows enhancing the variability space of images by synthetically generating augmentations of anatomical geometries and textures. Hence, variability can be specifically prescribed and controlled, for example, ranges of absolute or relative size, absolute or relative echogenicity, or texton granulometry, based on clinical needs.
It allows enriching the data with rare and underrepresented examples, contributing to the trained model's generalization and performance on such rare cases.

To address this need, we propose mask and image editing approaches.
Text-guided mask editing that controls geometrical properties is made possible with computer vision-based building blocks. For instance, enlarging an object in a semantic mask can be performed by the following operations:
\begin{enumerate*}[label=\alph*.]
    \item Filter the relevant object's semantic mask color.
    \item Find the minimal enclosing bounding box of the object.
    \item Scale in the x-axis and/or y-axis of the cropped bounding box.
    \item Blend the modified object into the original image while replacing the original object.
\end{enumerate*}
Similarly, operations such as object translation and rotation can be defined. We employ a pre-trained Large language model (LLM) \cite{touvron2023llamaopenefficientfoundation, chowdhery2022palmscalinglanguagemodeling} and prompt it to map a textual prompt to a sequence of classical computer vision operations, enabling text-based mask editing.

Importantly, the proposed method is general and can be applied to any mask to achieve text-guided mask editing, while the prompt has to be adapted to different tasks.

To control the texture properties of the image-guided texture, we utilize the Poisson Image Editing technique \cite{perez2023poisson}. This enables us to fuse a source image with a corresponding object provided by the semantic mask, and a target image from a specific device or campus.

\subsection{Experimental setup}\label{subsec:experimental}
\subsubsection{Study population and dataset}\label{data_section}
The dataset comprises 388 ultrasound images from 90 patients (48 males, and 42 females, with different BMI distribution).
Images included regions of the supraspinatus tendon, infraspinatus tendon, and subscapularis, from both the short-axis and long-axis views - all from the shoulder joint. 
The images were scanned using Logic S7 and E10 ultrasound systems from multiple clinical sites. To avoid training bias, the images were manually split to build an unbiased training set ($310$ images) and test set ($78$ images).
These images were manually annotated by trained personnel and verified by a clinical expert to distinguish between MSK anatomies and pathologies. 
Each pixel was assigned to one of eight classes, where Table \ref{Table:data_semanticlabelling} summarizes the semantic labeling statistics.
All images are resized to a constant resolution of 1535x796 and 1044x646 pixels.
We used this dataset to model the generation of synthetic ultrasound images.

\begin{table}[b]
\caption{Semantic labeling statistics.}
{
\centering
\begin{tabular}{l c c c}
 \hline
Class & \makecell{\# of polygons \\ (\% of total)\(^a\)} & \makecell{Total Area \\(pixels)\(^b\)} & \makecell{Mean fraction out\\ of total patch area\(^c\)} \\
\hline
\hline
   Bone & 288 (74.22\%) & 2.75M & 2.66\% \\
   Tendon & 388 (100\%)  & 10.64M & 10.31\% \\
   Muscle & 388 (100\%) & 28.13M & 27.26\% \\
   DITF & 179 (46.13\%) & 1.1M & 1.07\% \\
   Calcification & 44 (0.11\%) & 0.044M & 0.043\% \\
   Bone irregularity & 121 (0.31\%) & 1.22M & 1.19\% \\
   Anisotropy & 121 (0.31\%) & 2.9M & 2.03\% \\
   Background & 388 (100\%) & 57.2M & 55.42\%\\
\hline
\label{Table:data_semanticlabelling}
\end{tabular}\par
}
{\raggedright \footnotesize $^a$, The total number and percentages of 512x512 pixel patches containing at least one instance of the indicated class.\par}
{\raggedright \footnotesize $^b$, The pixels sum of the polygons area of the indicated class. \par}
{\raggedright \footnotesize $^c$, The mean percentage of pixels classified as the indicated class per total patch area.\par}
{\raggedright \footnotesize DITF, Discontinuity in tendon fiber; M, Million.\par}
\label{table:dataset_mask_distribution}
\end{table}

\subsubsection{Training procedure}\label{training_pipeline}
Our training procedure consists of (i) Fine-tuning of the mask generator network, StyleGAN, initialized with the weights of a pretrained encoder on BRECAHAD dataset (Section \ref{CSG_mask}), (ii) Using a Style loss \cite{gatys2015neural} and a frozen encoder to extract for each image the style features to find the closest non-equal descriptor in terms on MSE (Section \ref{CSG_context}), and (iii) Fine-tuning the image generator network, LDM pretrained on LAION-400M \cite{schuhmann2021laion} (Section \ref{CSG_image}). Here, the implementation details are as follows:
\begin{itemize}
    \item Mask generator - The StyleGanV2 \cite{karras2020training} architecture is finetuned with the AdamW \cite{loshchilov2017decoupled} optimizer with lr = \(1e^{-4}\). Batch size = $32$. 
    \item Image generator - The Latent diffusion model \cite{rombach2022high} is fine-tuned with the AdamW lr = \(3e^{-4}\). Batch sizes = $16$ (accumulate over 4). 
\end{itemize}
Experiments were trained with PyTorch \cite{NEURIPS2019_9015} on a single node NVIDIA A6000 GPUs with 32-bit floating point precision.  The models are fine-tuned with the AdamW \cite{loshchilov2017decoupled} optimizer with default parameters of \(\beta_{1} = 0.9, \beta_{2} = 0.999\).

\subsubsection{Evaluation metrics}\label{metrics}
Following previous work, we evaluated denovo musculoskeletal ultrasound image quality by image quality metrics and segmentation scores.
\begin{itemize}
    \item Image quality metrics: Kolmogorov–Smirnov Test (KST) \cite{KS_metric}, Kullback–Leibler Divergence (KLD) \cite{KL_metric} and Fréchet inception distance (FID) \cite{heusel2017gans} calculated between synthetic and real test images using a pre-trained Inception-v3 model \cite{szegedy2016rethinking}.
    \item Semantic segmentation metrics include the Dice Similarity Coefficient (DSC), Intersection over Union (IoU), Positive predictive value (PPV), True positive rate (TPR), and $F_1$-score (\ref{eq:dice} - \ref{eq:f1}).
    \begin{equation}
    \text{DSC} = \frac{2 \rvert P \cap GT \lvert }{\rvert P \lvert + \rvert GT \lvert} = \frac{2TP}{2TP + FP + FN} \label{eq:dice}
    \end{equation}
    \begin{equation}
        \text{IoU} = \frac{\rvert P \cap GT \lvert}{\rvert P \cup GT \lvert} = \frac{TP}{TP + FP + FN} \label{eq:iou}
    \end{equation}
    \begin{equation}
        \text{PPV} = \frac{TP}{TP + FP} \label{eq:precision}
    \end{equation}
    \begin{equation}
        \text{TPR} = \frac{TP}{TP + FN} \label{eq:recall}
    \end{equation}
    \begin{equation}
        \text{$F_1$-score} = \frac{2}{\frac{1}{\text{PPV}} + \frac{1}{\text{TPR}} }\label{eq:f1}
    \end{equation}
    
    The classification metrics, $TP$, $FP$, $FN$, (and $TN$) denote the true positive, false positive, false negative, (and true negative) of the areas of each patch, respectively.
    
    The DSC \cite{Sørensen-Dice} measures the overlapping pixels between prediction $\mathrm{P}$ and ground truth $\mathrm{GT}$ masks pixels.
    The IoU \cite{girshick2015fast} is the ratio between the area of intersection and the area of the union between the predicted $\mathrm{P}$ and ground truth $\mathrm{GT}$ masks pixels.
\end{itemize}

\section{RESULTS}\label{results}

\begin{figure*}[t!]
 \centering 
 \begin{subfigure}[b]{0.5\textwidth}
 \centering
\includegraphics[width=\textwidth]{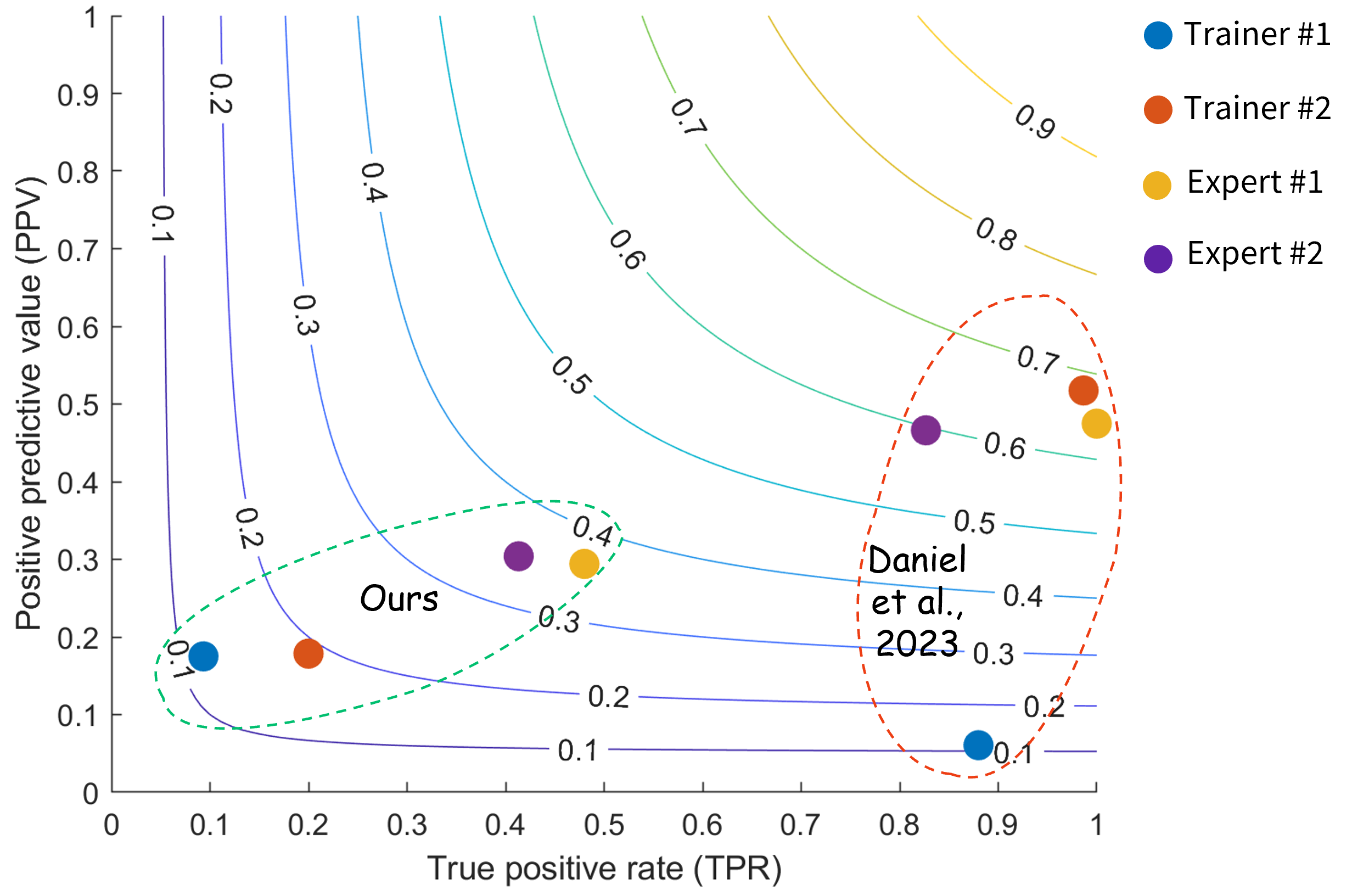}
        \caption{A scatter-plot of the CSG (ours) (green circle dash) and \cite{daniel2023between} (red circle dash) models Turing test results. The precision and recall values are derived from the predictions of the different models and human (filled circle point) with respect to the same test. The isolines lines are constant $F_1$-score values.}
        \label{f:f3_a}
    \end{subfigure}%
    \hspace{6pt}
    \begin{subfigure}[b]{0.48\textwidth}
    \centering
 \includegraphics[width=0.65\textwidth]{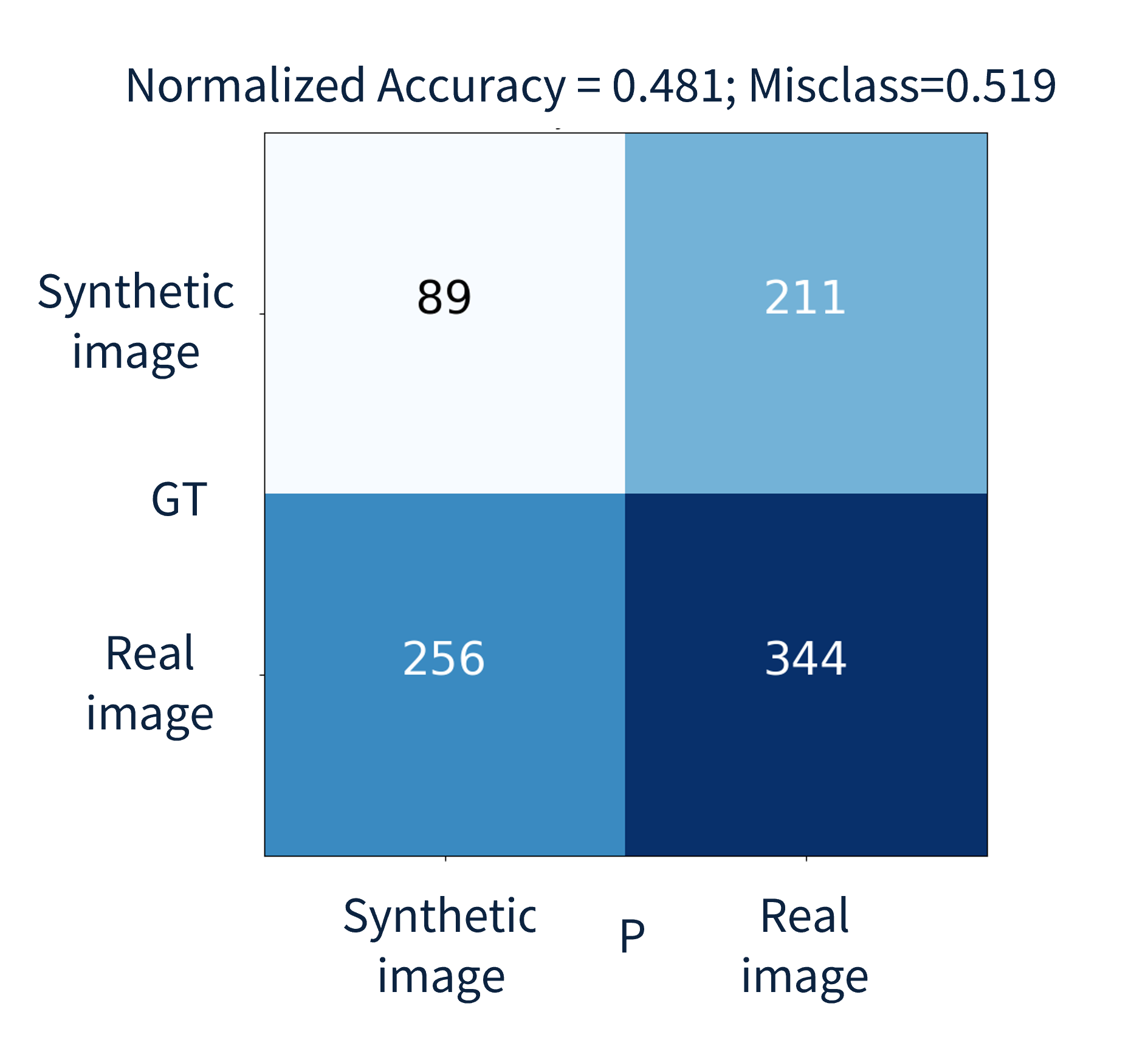}
        \caption{A confusion matrix result of the Turing test conducted with synthetic images generated by the CSG method.}
        \label{f:f3_b}
    \includegraphics[width=\textwidth]{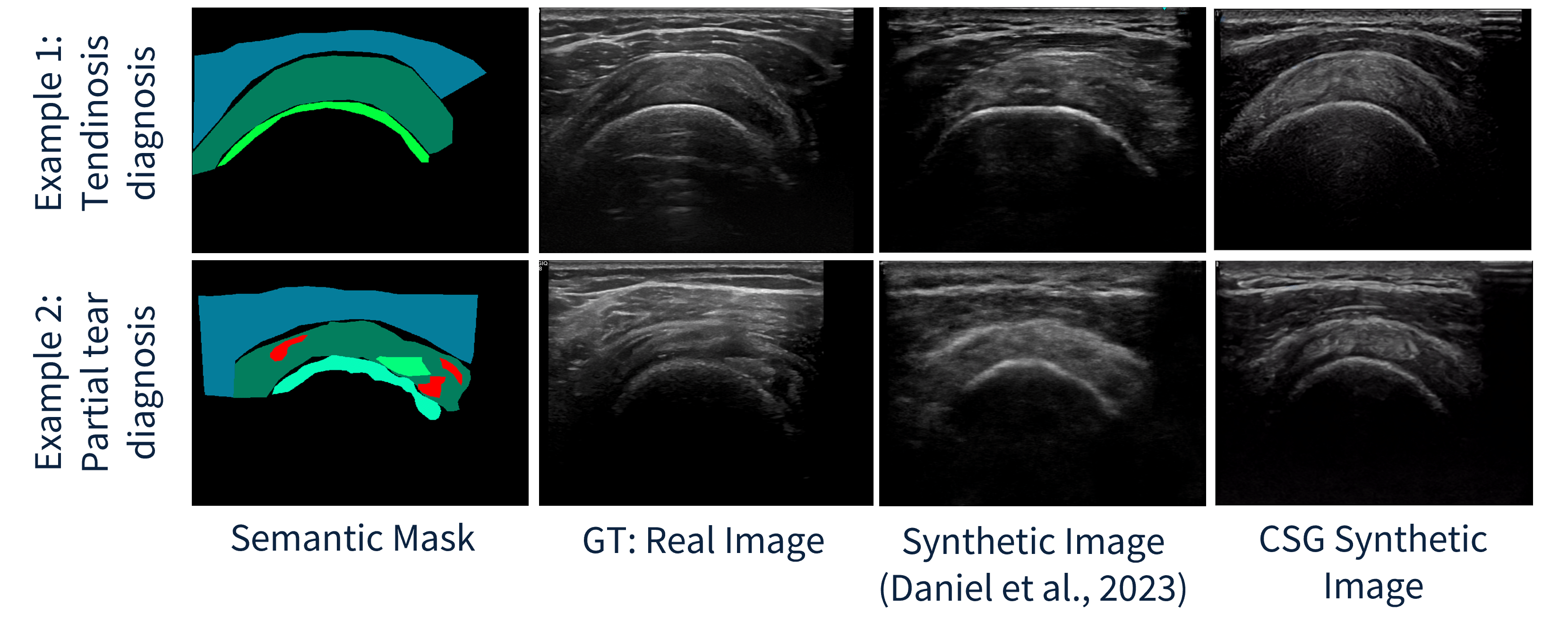}
        \caption{Synthetic images examples of our method compared to \cite{daniel2023between}.}
        \label{f:f3_c}
    \end{subfigure}
 \caption{Turing test results of the MSK synthetic images and a comparison of generated image examples.}
  \label{f:f3_new}
\end{figure*}

To validate the gain with synthetic images, we compare our results with other algorithms using a gold standard three-fold validation protocol: algorithmic improvement, image quality, and a Turing test. We further validate the ability of the method to extend the variability space.

\subsection{Algorithmic improvement}
We validated that using synthetic data in addition to real data improves the performance of a semantic segmentation model in the ultrasound imaging domain, and the MSK pathology field in particular. Furthermore, we evaluated the performance of two synthetic image generation algorithms: \cite{daniel2023between}, and ours (CSG). 
Thus, we trained 3 segmentation models \cite{Chen2017rethinking} on MSK ultrasound pathology dataset, and evaluated each with the DSC. 
The experimental dataset includes 388 real images from 124 subjects and 300 synthetic images that were generated either by the baseline ITM or the CSG method, where 80\% of which are used for training, and the remaining 20\% are used for validation. The test set comprises 40 images.
Table \ref{tab:mask_performance_analysis} shows that we improved by a factor of 26 and 23 on the test samples of all MSK classes and DITF respectively, compared to a controlled model, which is trained only on real images. Furthermore, CSG demonstrates an improvement over the baseline ITM \cite{daniel2023between}.

\begin{table}[!htbp]
    \caption{Quantitative segmentation test results of the mean DSC for different approaches on Musculoskeletal ultrasound images. Up arrow symbol indicates higher is better.}
    \centering
        {
        \begin{tabular}{l c c}
        \hline
             Methods & \makecell{mean DCS $\uparrow$ \\(all classes)} & \makecell{mean DCS $\uparrow$ \\(DITF)}\\
			\hline
			\hline
                  Real U/S images (control) \cite{Chen2017rethinking} & 0.46 & 0.39 \\
     		   Real U/S images w/ Baseline ITM \cite{daniel2023between}  & 0.56 & 0.44 \\
        	   Real U/S images w/ CSG (Ours) & \textbf{0.58} & \textbf{0.48}\\		
            \hline
		\end{tabular}\par
        }
    \label{tab:mask_performance_analysis}
\end{table}

\subsection{Metric validation}
The similarity between synthetic images and real images is estimated by applying the KST, KLD, and FID metrics.
Results in Table \ref{tab:t3} show that for MSK US dataset, CSG synthetic ultrasound images have the least distance to the real ones compared to \cite{daniel2023between}, in all metrics. Particularly, the FID score of CSG is lower (closer to the real distribution images) than the baseline ITM \cite{daniel2023between} by factor of 3.09.
Hence, in the context of the similarity, it can be observed that 
diffusion-based models generate de novo samples of higher fidelity than image translation-based models.

Following the manifold hypothesis, we postulate that high-fidelity synthetic images lie close to real images in the embedding space. To evaluate this hypothesis, the scattering of real and synthetic image distributions in the baseline's and our model's embedding space are used to construct their contours, by finding the convex hull of the images' representation t-SNE projection. Table \ref{tab:t3} shows an analysis of the scattering of real and synthetic distributions with various segmentation metrics. The results show that the overlap of synthetic CSG images with the real image distribution is greater compared to the baseline ITM \cite{daniel2023between}. 

\begin{table}[!htbp]
	\centering
    \caption{Objective quality assessment of de novo Musculoskeletal ultrasound images. $\downarrow$ denotes lower is better, and vice versa.}
		\begin{center}
		\begin{tabular}{l c c c c c c c c}
			\hline
		         & \multicolumn{3}{c}{Image quality metrics}
		         & \multicolumn{5}{c}{Contour segmentation metrics}\\
		         \cmidrule{2-4}
                  \cmidrule{6-9}
                 Methods & KST $\downarrow$ & KLD $\downarrow$ & FID $\downarrow$ && IoU $\uparrow$ & PPV $\uparrow$ & TPR $\uparrow$ & $F_1$-score $\uparrow$ \\
			\hline
			\hline
     		   Baseline ITM \cite{daniel2023between} & 0.072 & 0.621 & 24.189 && 0.270 & \textbf{0.437} & 0.414 & 0.425\\
        	   CSG (Ours) & \textbf{0.025} & \textbf{0.065} & \textbf{7.816} && \textbf{0.316} & 0.376 & \textbf{0.665} & \textbf{0.480}\\	
			\hline
		\end{tabular}
	\end{center}
	\label{tab:t3}
\end{table}

\subsection{Turing test}
Visual Turing tests were evaluated by trained personnel and clinical experts to judge the authenticity of ultrasound MSK images generated by our CSG method in comparison to those generated by the method of \cite{daniel2023between}.
Three sets of real and synthetic images were presented, each set comprised 100 images and had a different composition of synthetic and real images ($50\%-50\%; 70\%-30\% ;30\%-70\%$). 

The inter-observer agreement of trained personnel and clinical experts was high on both synthetic images ($82.6\%, 68\%$ respectively) and on real images ($57\%, 62.6\%$ respectively).

Fig.~\ref{f:f3_a} shows a scatter plot of the precision and recall values and $F_1$ scores as isolines of both methods. The results show that the synthetic images generated by the CSG method are less distinguishable from real images, than those generated by the method of \cite{daniel2023between}. A confusion matrix calculated based on the classification of images generated by the CSG method reveal a mean TPR of $0.2967$, mean PPV of $0.24$ and $F_1$ of 0.238, further attesting the authenticity of generated images.

A comparison of generated image examples shows that the synthetic images of our method not only match the geometric constraint of the semantic input mask, but also with context guidance, are of higher fidelity and resolution that better resemble the ground truth images in comparison to the method of \cite{daniel2023between} (Fig. \ref{f:f3_c}). 

\subsection{Extending the variability space}
\begin{figure}[!ht]
\centering
    \begin{subfigure}[t]{0.6\textwidth}
        \centering
        \includegraphics[width=0.7\linewidth]{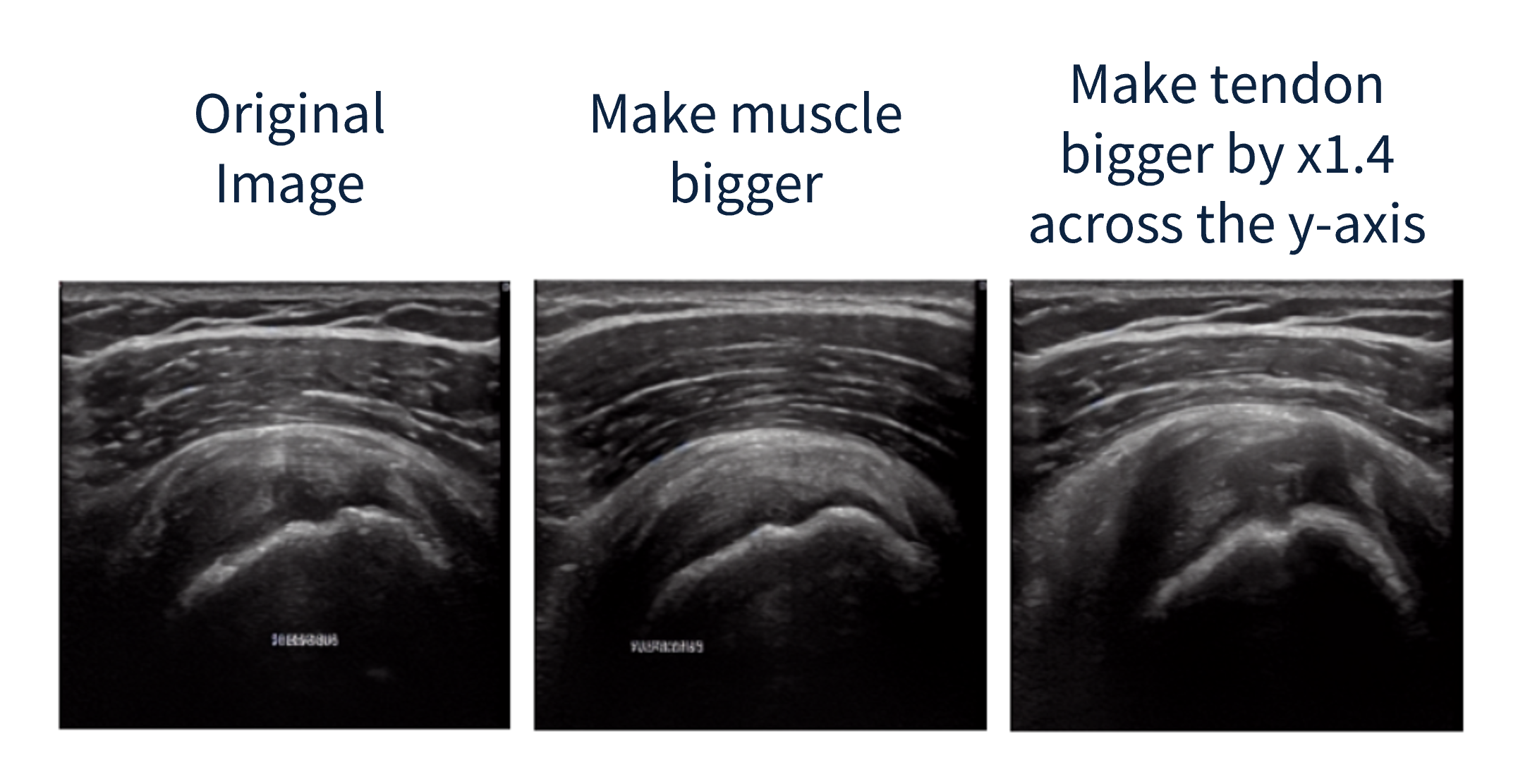}
    \end{subfigure}
    \begin{subfigure}[t]{\columnwidth}
        \centering
        \includegraphics[width=0.7\linewidth]{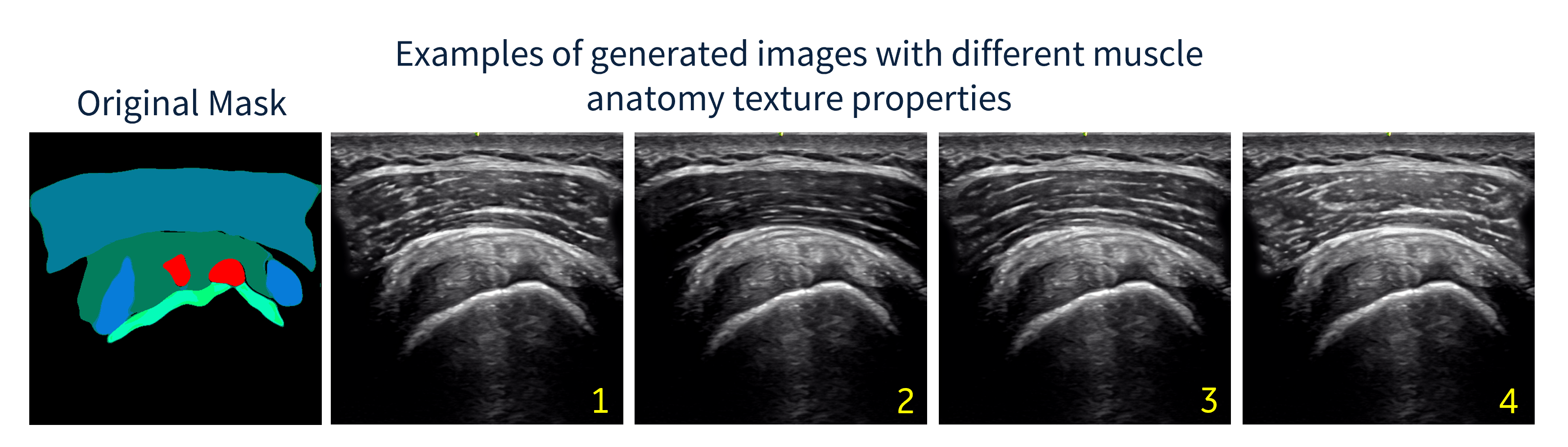}
    \end{subfigure}  
    \caption{Visualization of our extending the variability space. Our method allows for controlling the generation of the synthetic images by constraining its geometrical semantic properties and guiding its textural context properties. Top row: Text-guided mask editing results. Bottom row: Image-guided local texture editing results.}
    \label{f:f5_temp}
\end{figure}

The proposed extension of the CSG method successfully achieves text-guided mask editing and image texture control. Fig.~\ref{f:f5_temp} demonstrated modifying the geometry of the mask object based on the text prompt, resulting in consistent changes in the corresponding edited image. Further demonstrated is the Poisson Image Editing technique \cite{perez2023poisson}, which effectively blends the texture of the muscle anatomy from one image with another, showing smooth integration of textures while maintaining image realism. These results illustrate the potential of the method to address clinical needs by generating synthetic augmentations of anatomical geometries and textures.

\section{CONCLUSIONS}\label{conclusion}
In this work, we introduced a method for generating high-resolution images based on anatomical structure (geometry) and context guidance (texture). The results of the three-fold validation protocal show that our method produces synthetic images that are of superior image quality and performance improvement of downstream tasks in ultrasound imaging, while being indistinguishable from real MSK images, even for experts.

In general, our approach provides a state-of-the-art solution to the demanding task synthetic ultrasound images generation, incorporating semantic and contextual guidance in a controllable way. Its extension allows further enhancing the variability of anatomical geometries and textures, which may increase the representation of rare examples in the data.
Lastly, it facilitates the progress of ultrasound diagnosis by enhancing the performance and robustness of AI models by producing synthetic images that are critical for many applications and systems.

\section{Compliance with Ethical Standards}
This research study was conducted retrospectively using human subject data internally collected by GE HealthCare. Data collection followed the principles outlined in the Declaration of Helsinki (1975, revised in 2000). All procedures performed in this study involving IRB have an approved data sharing agreement.


\bibliographystyle{unsrt}

\end{document}